\pdfoutput=1

\PassOptionsToPackage{dvipsnames}{xcolor}
\documentclass[11pt]{article}
\usepackage[]{acl}

\usepackage{times}
\usepackage{latexsym}

\usepackage[T1]{fontenc}

\usepackage[utf8]{inputenc}

\usepackage{color}
\usepackage{xcolor}
\usepackage{colortbl}
\usepackage[normalem]{ulem}

\usepackage{xurl}

\usepackage{tabularx}
\usepackage{tabu}
\usepackage{booktabs}
\usepackage{multirow}

\usepackage{graphicx}
\usepackage{float}
\usepackage{placeins}
\usepackage{subfigure}
\usepackage{amssymb}

\usepackage{hyperref}
\usepackage{tablefootnote}
\usepackage{xspace}
\usepackage{tcolorbox}
\usepackage{fancyvrb}

\usepackage{float}
\usepackage{amsmath}
\usepackage{bm}
\usepackage{algorithmicx}
\usepackage{algorithm}
\usepackage{algpseudocode}

\usepackage{arydshln}

\usepackage{amssymb}
\usepackage{pifont}

\usepackage{enumitem}

\usepackage{microtype}

\makeatletter
\long\def\@makefntext#1{\noindent$^{\@thefnmark}$\hspace{0.25em}#1}
\makeatother

\newcommand{\hflogo}{\raisebox{-1.5pt}{\includegraphics[height=1.05em]{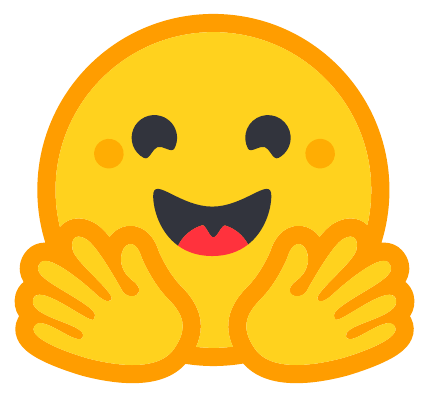}}\xspace}
\newcommand{\ghlogo}{\raisebox{-1.5pt}{\includegraphics[height=1.05em]{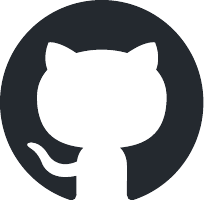}}\xspace}

\definecolor{YaleBlue}{RGB}{0, 53, 107}
\definecolor{NYUPurple}{RGB}{87, 6, 140}
\definecolor{NTUGreen}{RGB}{0, 150, 57}
\definecolor{UCASRed}{RGB}{200, 16, 46}
\definecolor{SummaryColGray}{gray}{0.93}

\newcommand{\Yale}{\hspace{.1em}^{\textcolor{YaleBlue}{\boldsymbol{Y}}}}
\newcommand{\NYU}{\hspace{.1em}^{\textcolor{NYUPurple}{\boldsymbol{N}}}}
\newcommand{\NTU}{\hspace{.1em}^{\textcolor{NTUGreen}{\boldsymbol{T}}}}

\title{GUI vs. CLI: Execution Bottlenecks in Screen-Only and \\ Skill-Mediated Computer-Use Agents}

\author{
Xiao Zhou$\NYU\Yale$\thanks{Equal contributions. Correspondence to: Yilun Zhao
(yilun.zhao@yale.edu), Chen Zhao (cz1285@nyu.edu).}\quad
Siyue Zhang$\NTU$\footnotemark[1]\quad
Yilun Zhao$\Yale$\quad
Jinbiao Wei$\Yale$ \\[3pt]
\textbf{
Tingyu Song$\Yale$\quad
Arman Cohan$\Yale$\quad
Chen Zhao$\NYU$} \\[8pt]
$\NYU$NYU Shanghai \quad
$\Yale$Yale NLP Lab \quad
$\NTU$Nanyang Technological University \\[7pt]
\begin{tabular}{clcl}
\hflogo & \href{https://huggingface.co/datasets/rebeccazzzz/gui-vs-cli}{Data \& Models} &
\ghlogo & \href{https://github.com/rebeccaz4/gui-vs-cli.git}{Code}
\end{tabular}

}

\begin{document}
\maketitle
\begin{abstract}

Computer-use agents can execute software tasks through either graphical interfaces or programmatic command interfaces, but existing evaluations confound interaction modality with differences in tasks, initial states, verifiers, and permitted actions. We introduce a matched execution-layer benchmark of 440 desktop tasks across 18 applications and 12 workflow categories, where screen-only GUI agents and skill-mediated CLI agents receive identical goals, states, and final-state verifiers while being restricted to modality-native actions. In this controlled setting, the strongest GUI agent reaches a 59.1\% full pass rate, outperforming the strongest original-skill CLI agent at 48.2\%; however, verifier-guided skill augmentation raises CLI success to 69.3\%, showing that much of the CLI deficit comes from incomplete skill coverage rather than model capability alone. These results suggest that GUI and CLI expose different execution bottlenecks: GUI agents are limited by reliable grounded interaction over long-horizon workflows, whereas CLI agents are limited by the coverage and scalability of their skill interfaces.

\end{abstract}

\section{Introduction}
\begin{figure*}[t]
\centering
\includegraphics[width=\textwidth]{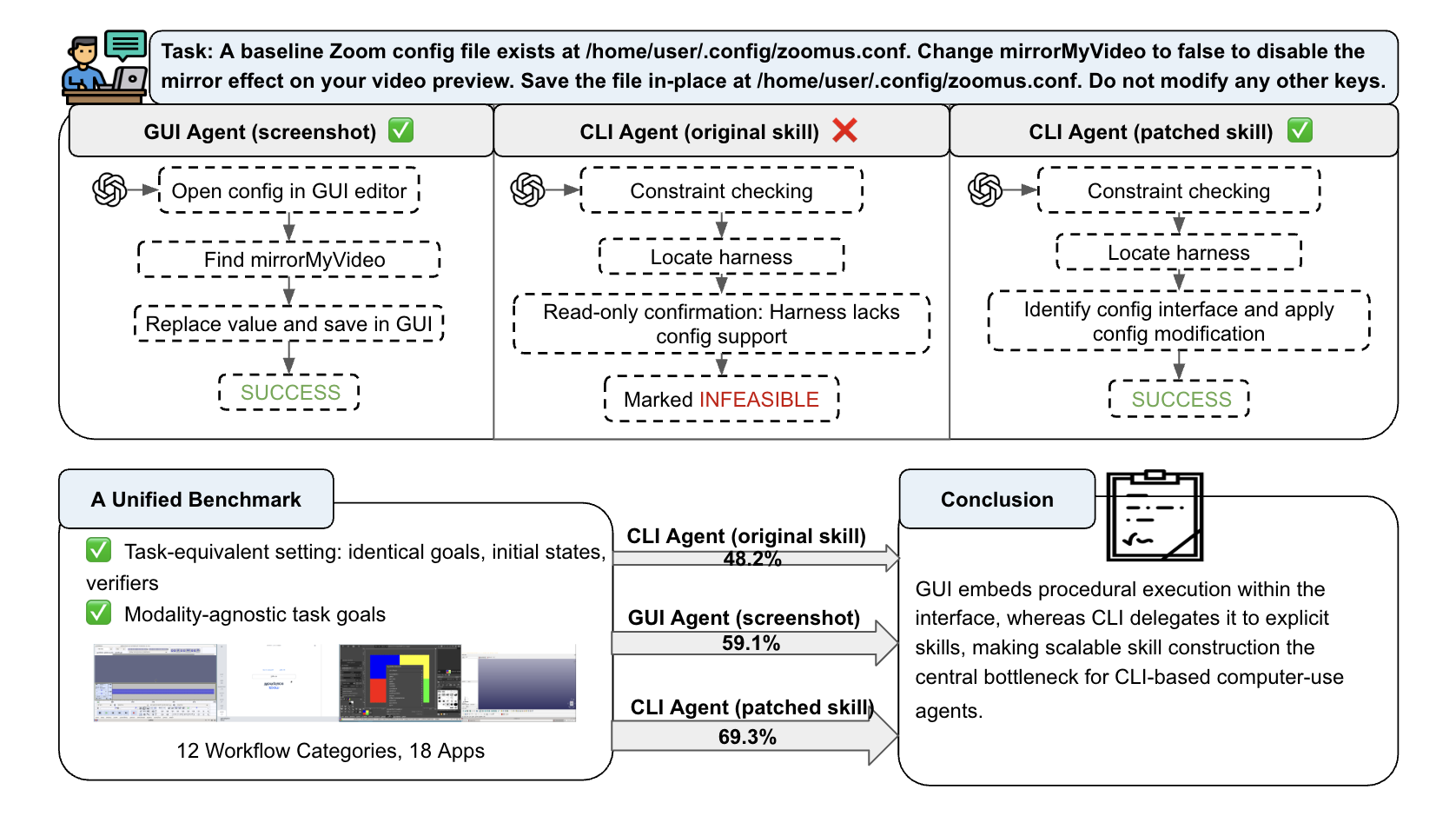}
\caption{\textbf{Overview.} A matched execution-layer benchmark of 440 desktop tasks compares GUI and skill-mediated CLI agents under identical goals, states, and verifiers. Results show that the observed CLI gap is strongly affected by skill coverage, while GUI embeds procedural execution within the interface.}
\label{fig:overview}
\end{figure*}

A desktop task such as renaming three tracks in Audacity, editing a slide deck, or constructing a multi-page diagram in drawio can be executed in two very different ways: by operating the visible application interface as a human would~\citep{xie2024osworld,guisurvey,wei2026opencomputerverifiablesoftwareworlds}, or by invoking programmatic skills that manipulate application state~\citep{anthropic2026sonnet46,openai2026gpt54,hkuds2026clianything,he2026skillsbench}.
Concretely, a GUI agent observes a screenshot of the visible application controls and acts through clicks, drags, typing, scrolling, and keyboard shortcuts on a sandboxed desktop~\citep{xie2024osworld}, while a skill-mediated CLI agent observes the application as a set of composable operations exposed by a curated skill layer~\citep{hkuds2026clianything,han2026sweskills} and acts by invoking those operations. To ensure a fair comparison, GUI agents are restricted to interacting with applications exclusively through graphical interfaces, whereas CLI agents are required to complete tasks using application-specific commands and skills, without relying on shortcuts such as directly modifying output files. The two settings therefore differ not only in surface mechanics, but in how application capability is represented and made actionable for the agent.

Yet existing evaluations cannot isolate the effect of interaction modality, because GUI-focused~\citep{zhou2023webarena,drouin2024workarena,rawles2024androidworld,xie2024osworld} and programmatic-agent~\citep{trivedi2024appworld,bechard2026terminal} benchmarks vary three confounded factors at once: the applications they target, the initial states and final-state verifiers they use, and the action spaces they permit.
When all three vary simultaneously, an outcome success cannot reveal whether performance differences come from the model, the task setup, or the interaction modality itself.
A controlled comparison must therefore hold the task goal, initial state, and final-state verifier fixed across modalities, while enforcing modality-native action spaces.

To address these confounds, we instantiate a matched execution-layer benchmark for GUI and skill-mediated CLI agents.
The benchmark contains 440 desktop tasks spanning 18 applications and 12 workflow categories, in which both modalities receive identical user goals, initial states, and executable final-state verifiers.
Each task is rewritten as a modality-neutral instruction describing only the desired outcome, and each agent is restricted to its modality-native action space (screen-only operations for GUI, the skill layer for CLI), so that task content, initial state, and verification are controlled while the native action interface is systematically varied across settings.

Across this benchmark, the strongest GUI agent (GPT-5.4) achieves a 59.1\% full pass rate while the strongest CLI agent (Codex GPT-5.5) reaches 48.2\% under the original CLI-Anything skill layer, with each modality favoring different workflows: GUI dominates where the application interface directly exposes the intended workflow, while CLI is competitive or stronger where the target state can be represented as structured artifacts.
We further diagnose how much of the CLI gap is explained by incomplete skill coverage.
Only 37.6\% of verifier checkpoints can be satisfied by the original skill interface.
In a verifier-guided patched-skill setting, where missing skill paths are repaired against verifier-observed requirements, CLI success rises to 69.3\%.
Because this repair process uses verifier information, we interpret the result as a coverage-controlled diagnostic upper bound rather than an out-of-the-box CLI baseline.
The remaining gaps in spreadsheet and web workflows suggest that modality-specific challenges persist beyond missing skill operations.
The two modalities further exhibit complementary failure modes, with GUI agents bottlenecked by visual grounding and long workflow execution, and CLI agents bottlenecked by skill coverage gaps and implicit-default reconstruction.
These results reframe the GUI versus CLI comparison as a question of where execution logic is engineered: into the application's visible interface, or into a constructed skill layer that defines the agent's available operations.

Our contributions include:
\begin{itemize}[noitemsep,topsep=0pt,leftmargin=*]
    \item A controlled execution-layer protocol that holds the task goal, initial state, and final-state verifier fixed across GUI and skill-mediated CLI agents while enforcing modality-native action spaces.
    \item A benchmark of 440 desktop tasks across 18 applications and 12 workflow categories, with modality-neutral instructions and executable checks over final states.
    \item A diagnostic analysis of skill coverage as a CLI bottleneck: original skills satisfy 37.6\% of verifier checkpoints, and a verifier-guided patched-skill setting raises CLI success from 48.2\% to 69.3\%, estimating the recoverable performance lost to skill-interface incompleteness.
    \item A taxonomy of complementary failure modes: visual grounding and long workflow execution for GUI; skill coverage gaps, implicit-default reconstruction, and unobservable application semantics for CLI.
\end{itemize}

\section{Related Work}

\paragraph{GUI Agents.}
Recent work has rapidly expanded both training methods~\citep{qin2025ui,wang2025opencuaopenfoundationscomputeruse,wei2026anchor,gan2026androidcoach} and benchmarks~\citep{zhou2023webarena,drouin2024workarena,xie2024osworld} for agents operating through graphical interfaces, spanning web and enterprise environments as well as full desktop operating systems.
Across these settings, perceptual grounding, long action chains, and recovery from layout changes recur as dominant failure modes, with oracle or manually grounded actions closing much of the performance gap~\citep{deng2023mind2web,zheng2024seeact,aghzal2026hierarchicalwebagents}.
Because these evaluations are entirely visual, they conflate task, model, and GUI-modality effects.

\paragraph{CLI Agents and Skills.}
Another line of work studies agents that act through programmatic interfaces rather than visual control.
Early benchmarks evaluate execution through APIs, command-line environments, or direct filesystem access~\citep{liu2023agentbench,trivedi2024appworld,li2023apibank,bechard2026terminal,merrill2026terminalbench,song2024beyond}, but these settings differ in what they expose: an unrestricted shell affords arbitrary code, while a curated skill layer constrains the agent to a fixed set of application-level operations.
Recent skill-based systems make this latter view explicit: CLI-Anything packages applications as reusable command-line harnesses~\citep{hkuds2026clianything}, and evaluations of curated skills~\citep{he2026skillsbench,han2026sweskills} find that the coverage and quality of the skill layer largely determine downstream success.
Our CLI setting follows this skill-mediated view rather than treating CLI access as unrestricted coding, which separates modality effects from general code-execution capability.

\section{Benchmark}
Our unified benchmark evaluates agents at the execution layer by comparing GUI and CLI interaction modalities under matched task goals, initial states, and final-state verification.

\subsection{Benchmark Scope and Composition}
\begin{figure}[t]
\centering
\includegraphics[width=0.88\linewidth]{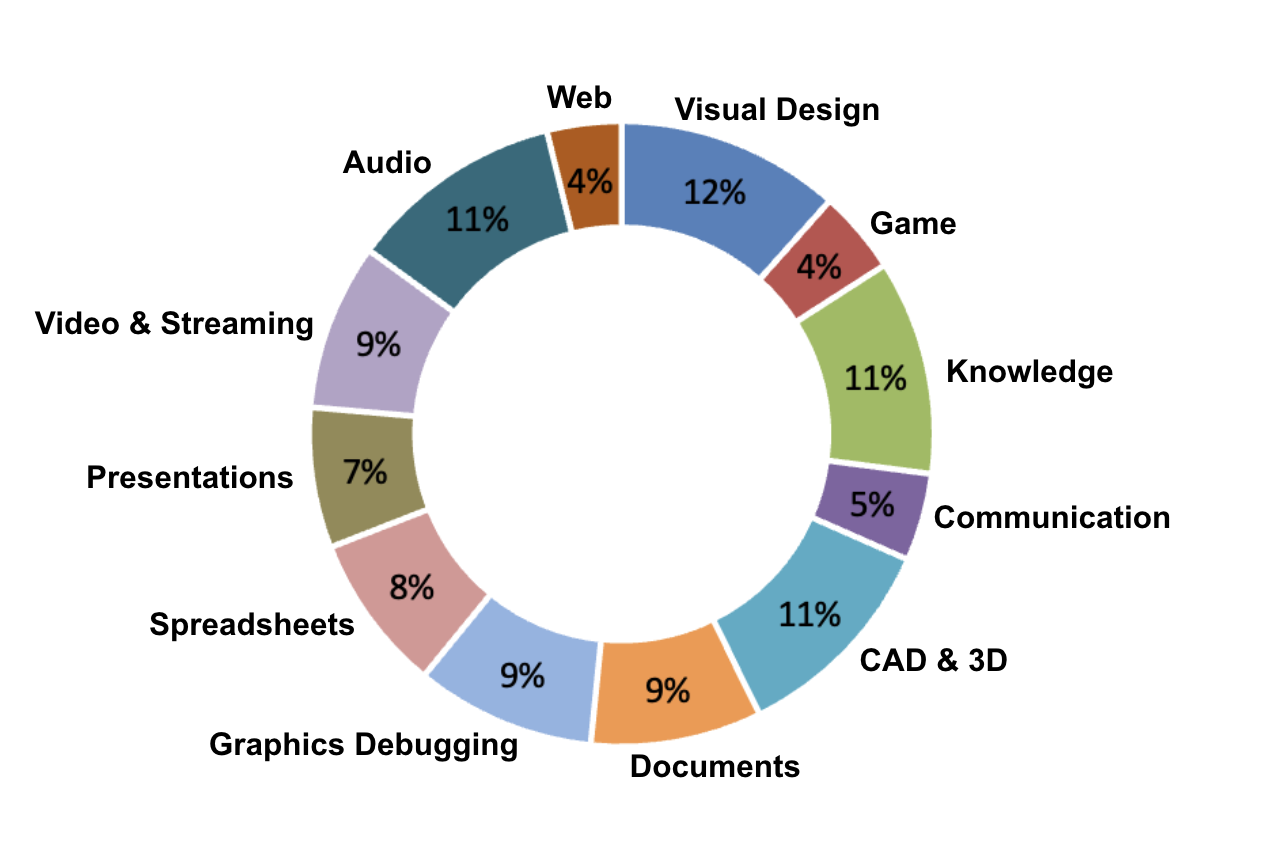}
\vspace{0.4em}

\small
\setlength{\tabcolsep}{5pt}
\begin{tabular}{ll}
\toprule
\textbf{Workflow} & \textbf{Applications} \\
\midrule
Visual Design & GIMP, Krita, draw.io \\
Audio & Audacity, MuseScore 3 \\
Knowledge & Obsidian, Zotero \\
CAD \& 3D & FreeCAD, CloudCompare \\
Graphics Debugging & RenderDoc \\
Documents & LibreOffice Writer \\
Video \& Streaming & Shotcut, OBS \\
Spreadsheets & LibreOffice Calc \\
Presentations & LibreOffice Impress \\
Communication & Zoom \\
Game & Godot 4 \\
Web & Chrome \\
\bottomrule
\end{tabular}

\caption{Benchmark composition across workflows and applications in the 440-task benchmark, spanning 18 applications across 12 workflow categories.}
\label{fig:domain-distribution}
\end{figure}

As shown in Figure~\ref{fig:domain-distribution}, the benchmark contains 440 tasks across 18 real-world applications and 12 workflow categories.
We organize tasks by workflow rather than by application alone, because modality effects often depend on the kind of state transformation required: some tasks rely on visual layout and interface navigation, while others expose structured artifacts or application-level operations.
This composition supports both aggregate comparison and per-workflow analysis of where GUI and CLI execution succeeds or fails.

\subsection{Benchmark Construction}
\begin{figure*}[t]
\centering
\includegraphics[width=\textwidth]{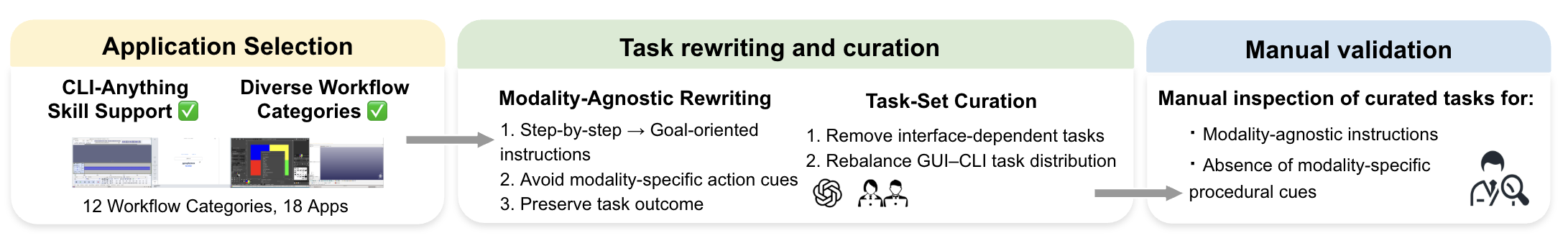}
\caption{Benchmark construction pipeline. We select applications with CLI-Anything skill support, rewrite GUI-oriented tasks into modality-agnostic task descriptions, curate the task set to reduce interface bias, and manually validate the resulting tasks.}
\label{fig:benchmark-pipeline}
\end{figure*}

We construct our benchmark through a three-stage pipeline, shown in Figure~\ref{fig:benchmark-pipeline}, that adapts existing verifiable desktop tasks into a matched GUI--CLI evaluation suite.

\paragraph{Stage I: Application and Task Selection.}
We start from tasks in OpenComputer~\citep{wei2026opencomputerverifiablesoftwareworlds} and select applications for which corresponding CLI-Anything~\citep{hkuds2026clianything} skills are available. This ensures that each selected application can be exercised through both visual interaction and programmatic execution.

\paragraph{Stage II: Task Rewriting and Curation.}
We rewrite the original step-by-step, GUI-oriented instructions into modality-agnostic task descriptions that specify the target outcome rather than a GUI-specific procedure and curate the resulting task set to balance the GUI--CLI execution distribution. Specifically, each rewritten instruction is given identically to both modalities and avoids modality-specific action sequences where possible. We remove tasks whose outcomes depend on one interface or lack a meaningful counterpart in the other modality. We also add tasks where needed to offset task-set imbalances.

\paragraph{Stage III: Manual Validation.}
We manually inspect curated tasks to ensure that each task targets execution-layer capability and can be given identically to both modalities. This validation checks both the task objective and its language: the task should be solvable through either modality under the benchmark constraints, and the instruction should avoid modality-specific procedural cues while specifying a precise target state that is evaluated by the same verifier in both settings.

\section{Experiment Setup}
We evaluate both modalities on the same benchmark, with identical instructions, initial states, and executable final-state verifiers.

\paragraph{Action Space.}
To isolate the effect of interaction modality rather than general coding ability, we restrict agents from using unrestricted programming or direct artifact manipulation to bypass the evaluated interface. We enforce this constraint through the agent prompts shown in Appendix~\ref{app:prompts}. In the GUI setting, agents are restricted to screen-mediated interaction with a sandboxed Ubuntu desktop: at each step, the agent receives a screenshot and may output only screen-level computer actions, such as clicking, dragging, typing, scrolling, and keyboard shortcuts. Direct code execution, shell commands, and programmatic filesystem or database edits are disallowed; task-result changes must be performed through visible GUI operations. In the CLI setting, agents are restricted to CLI-Anything skills. The shell may be used to discover and invoke relevant skills, manage execution, and perform read-only inspection or verification, but task-result changes must be made through skill-provided workflows or application-level commands rather than direct edits to files, databases, or artifacts.

\paragraph{Models.}
For GUI agents, we evaluate GPT-5.4~\citep{openai2026gpt54}, Claude-Sonnet-4.6~\citep{anthropic2026sonnet46}, Claude-Opus-4.7~\citep{anthropic2026opus47}, EvoCUA-32B-20260105~\citep{xue2026evocua}, Qwen3.5-27B~\citep{qwen2026qwen3527b}, and Kimi-K2.6~\citep{moonshot2026kimi26}. For CLI agents, we evaluate two Codex variants, GPT-5.4 and GPT-5.5~\citep{openai2026gpt54,openai2026gpt55}, and two Claude Code variants, Claude-Sonnet-4.6 and Claude-Opus-4.7~\citep{anthropic2026sonnet46,anthropic2026opus47}, as CLI baselines.

\paragraph{Evaluation Metrics.}
Our evaluation follows an execution-based protocol: task success is determined by the final application state rather than by matching a reference action trajectory. Each task contains executable verifier checkpoints over task-relevant application state and artifacts. We report full pass rate, requiring all verifier checks for a task to pass, and average task time as an execution-cost measure for each setting. 

\section{Main Results}

\begin{table*}[t!]
\centering
\small
\setlength{\tabcolsep}{3.2pt}
\resizebox{\textwidth}{!}{
\begin{tabular}{ll*{12}{c}>{\columncolor{SummaryColGray}}c>{\columncolor{SummaryColGray}}c}
\toprule
\textbf{Modality} & \textbf{Model}
& \textbf{Vis.} & \textbf{Aud.} & \textbf{Know.} & \textbf{CAD.} & \textbf{Gra.} & \textbf{Doc.} & \textbf{Vid.} & \textbf{Sheet.} & \textbf{Slide.} & \textbf{Comm.} & \textbf{Game} & \textbf{Web}
& \textbf{Avg.} & \textbf{Time (s)} \\
\midrule
\multirow{6}{*}{\textbf{GUI}}
& GPT-5.4            & \textcolor{black}{\underline{47.1}} & \textcolor{black}{\underline{73.5}} & \textcolor{black}{\textbf{42.9}} & \textcolor{black}{46.9} & \textcolor{black}{51.2} & \textcolor{black}{51.3} & \textcolor{black}{\textbf{60.5}} & \textcolor{black}{\underline{61.1}} & \textcolor{black}{\textbf{78.1}} & \textcolor{black}{\textbf{70.0}} & \textcolor{black}{84.2} & \textcolor{black}{\textbf{88.2}} & \textcolor{black}{\textbf{59.1}} & \textcolor{black}{455.8} \\
& Claude Sonnet 4.6  & \textcolor{black}{43.1} & \textcolor{black}{69.4} & \textcolor{black}{\underline{34.7}} & \textcolor{black}{\underline{63.3}} & \textcolor{black}{\underline{56.1}} & \textcolor{black}{25.6} & \textcolor{black}{50.0} & \textcolor{black}{33.3} & \textcolor{black}{34.4} & \textcolor{black}{45.0} & \textcolor{black}{84.2} & \textcolor{black}{70.6} & \textcolor{black}{49.1} & \textcolor{black}{245.4} \\
& Claude Opus 4.7    & \textcolor{black}{37.3} & \textcolor{black}{\textbf{81.6}} & \textcolor{black}{\underline{34.7}} & \textcolor{black}{59.2} & \textcolor{black}{\textbf{70.7}} & \textcolor{black}{\underline{53.9}} & \textcolor{black}{\underline{55.3}} & \textcolor{black}{\textbf{63.9}} & \textcolor{black}{\underline{56.3}} & \textcolor{black}{15.0} & \textcolor{black}{73.7} & \textcolor{black}{70.6} & \textcolor{black}{\underline{55.9}} & \textcolor{black}{346.4} \\
& EvoCUA-32B         & \textcolor{black}{23.5} & \textcolor{black}{34.7} & \textcolor{black}{28.6} & \textcolor{black}{16.3} & \textcolor{black}{22.0} & \textcolor{black}{15.4} & \textcolor{black}{31.6} & \textcolor{black}{5.6} & \textcolor{black}{15.6} & \textcolor{black}{35.0} & \textcolor{black}{21.1} & \textcolor{black}{52.9} & \textcolor{black}{23.9} & \textcolor{black}{254.4} \\
& Qwen3.5-27B        & \textcolor{black}{35.3} & \textcolor{black}{24.5} & \textcolor{black}{8.2} & \textcolor{black}{14.3} & \textcolor{black}{2.4} & \textcolor{black}{5.1} & \textcolor{black}{26.3} & \textcolor{black}{11.1} & \textcolor{black}{18.8} & \textcolor{black}{10.0} & \textcolor{black}{42.1} & \textcolor{black}{64.7} & \textcolor{black}{19.3} & \textcolor{black}{1306.8} \\
& Kimi-K2.6          & \textcolor{black}{41.2} & \textcolor{black}{51.0} & \textcolor{black}{32.7} & \textcolor{black}{51.0} & \textcolor{black}{36.6} & \textcolor{black}{15.4} & \textcolor{black}{44.7} & \textcolor{black}{16.7} & \textcolor{black}{21.9} & \textcolor{black}{25.0} & \textcolor{black}{68.4} & \textcolor{black}{\underline{82.4}} & \textcolor{black}{38.6} & \textcolor{black}{1421.2} \\
\midrule
\multirow{4}{*}{\textbf{CLI}}
& Codex GPT-5.4      & \textcolor{black}{\underline{47.1}} & \textcolor{black}{20.4} & \textcolor{black}{8.2} & \textcolor{black}{61.2} & \textcolor{black}{4.9} & \textcolor{black}{5.1} & \textcolor{black}{36.8} & \textcolor{black}{0.0} & \textcolor{black}{9.4} & \textcolor{black}{5.0} & \textcolor{black}{\underline{89.5}} & \textcolor{black}{0.0} & \textcolor{black}{24.3} & \textcolor{black}{254.4} \\
& Codex GPT-5.5      & \textcolor{black}{\textbf{54.9}} & \textcolor{black}{42.9} & \textcolor{black}{22.4} & \textcolor{black}{\textbf{67.3}} & \textcolor{black}{9.8} & \textcolor{black}{\textbf{64.1}} & \textcolor{black}{\textbf{60.5}} & \textcolor{black}{47.2} & \textcolor{black}{46.9} & \textcolor{black}{\underline{50.0}} & \textcolor{black}{\textbf{100.0}} & \textcolor{black}{35.3} & \textcolor{black}{48.2} & \textcolor{black}{\textbf{188.1}} \\
& Claude Code Sonnet 4.6 & \textcolor{black}{\underline{47.1}} & \textcolor{black}{30.6} & \textcolor{black}{4.1} & \textcolor{black}{40.8} & \textcolor{black}{2.4} & \textcolor{black}{10.3} & \textcolor{black}{39.5} & \textcolor{black}{5.6} & \textcolor{black}{15.6} & \textcolor{black}{0.0} & \textcolor{black}{\textbf{100.0}} & \textcolor{black}{23.5} & \textcolor{black}{25.2} & \textcolor{black}{\underline{208.6}} \\
& Claude Code Opus 4.7   & \textcolor{black}{41.2} & \textcolor{black}{20.4} & \textcolor{black}{8.2} & \textcolor{black}{36.7} & \textcolor{black}{17.1} & \textcolor{black}{12.8} & \textcolor{black}{39.5} & \textcolor{black}{5.6} & \textcolor{black}{15.6} & \textcolor{black}{0.0} & \textcolor{black}{\textbf{100.0}} & \textcolor{black}{5.9} & \textcolor{black}{24.3} & \textcolor{black}{248.8} \\
\bottomrule
\end{tabular}
}
\caption{The performance of GUI and CLI agents across workflow categories. We report full pass rate for all workflow categories: Visual Design (Vis.), Audio (Aud.), Knowledge (Know.), CAD \& 3D (CAD.), Graphics Debugging (Gra.), Documents (Doc.), Video \& Streaming (Vid.), Spreadsheets (Sheet.), Presentations (Slide.), Communication (Comm.), Game, and Web. Avg. denotes the overall full pass rate across 440 tasks. Time (s) reports the average wall-clock runtime per task in seconds. The best score in each column is highlighted in \textbf{bold}, and the second best is \underline{underlined}. For Time, lower is better.}
\label{tab:domain-results}
\end{table*}

\paragraph{Interaction Modality Can Compensate for Model Limitations.} As shown in Table~\ref{tab:domain-results}, the strongest GUI agent, GPT-5.4, achieves a 59.1\% full pass rate, followed by Claude Opus 4.7 at 55.9\%, while the strongest original skill-mediated CLI agent, Codex GPT-5.5, reaches 48.2\%. Since tasks, initial states, and verifiers are held constant across modalities, these differences reflect the combined effects of model capability and interaction design rather than task variation. Notably, GPT-5.4 operating through a GUI interface substantially outperforms Codex GPT-5.5 operating through a skill-mediated CLI, despite the latter employing a stronger underlying model. This result suggests that interaction design can narrow, and sometimes reverse, performance gaps between underlying models.

\paragraph{Interaction Modality Affects Robustness Across Diverse Workflow Categories.}

GUI agents show more stable performance across workflow categories, with GPT-5.4 ranging from 42.9\% to 88.2\%. In contrast, a CLI agent such as Codex GPT-5.5 ranges from 9.8\% to 100.0\%. The results suggest that GUI interaction benefits from execution structure embedded within application interfaces, which provides comparatively stable workflow guidance across applications. By comparison, CLI execution depends more heavily on the availability and reliability of workflow-specific skill abstractions, making performance more sensitive to skill coverage differences.

\paragraph{Different Modalities Favor Different Workflow Structures.}
The per-workflow results show that modality advantages are not uniform across the benchmark. GUI agents perform better on workflows where the human-facing interface directly exposes the intended operation sequence, such as Audio, Presentations, Communication, and Web tasks. These tasks often require interacting with visible controls, menus, timelines, slides, or browser state, where the GUI provides the native workflow representation. CLI agents are competitive or stronger in workflows where the target state can be represented as structured artifacts or application-level operations, including Visual Design, CAD \& 3D, Documents, Video \& Streaming, and Game tasks. The Visual Design result is especially informative: although the domain appears visually oriented, many tasks involve structured artifact properties, such as pages, shapes, labels, and connectors, that can be manipulated more directly through programmatic operations than through screen-level placement. Appendix~\ref{app:visual-design-example} provides an example. Thus, the modality advantage is tied less to the application category itself than to the representation each interface provides for execution.

\section{Error Analysis}

To understand why the two modalities fail in different ways, we further inspect failed trajectories and verifier outputs from the strongest GUI and CLI agents. Rather than treating failures only as aggregate labels, we use representative failed tasks to identify recurring mechanisms, and then generalize these mechanisms into an error taxonomy. Additional grounded trajectory-level examples for each taxonomy category are provided in Appendix~\ref{app:error-taxonomy}.

\begin{figure}[t]
    \centering
    \includegraphics[width=\linewidth]{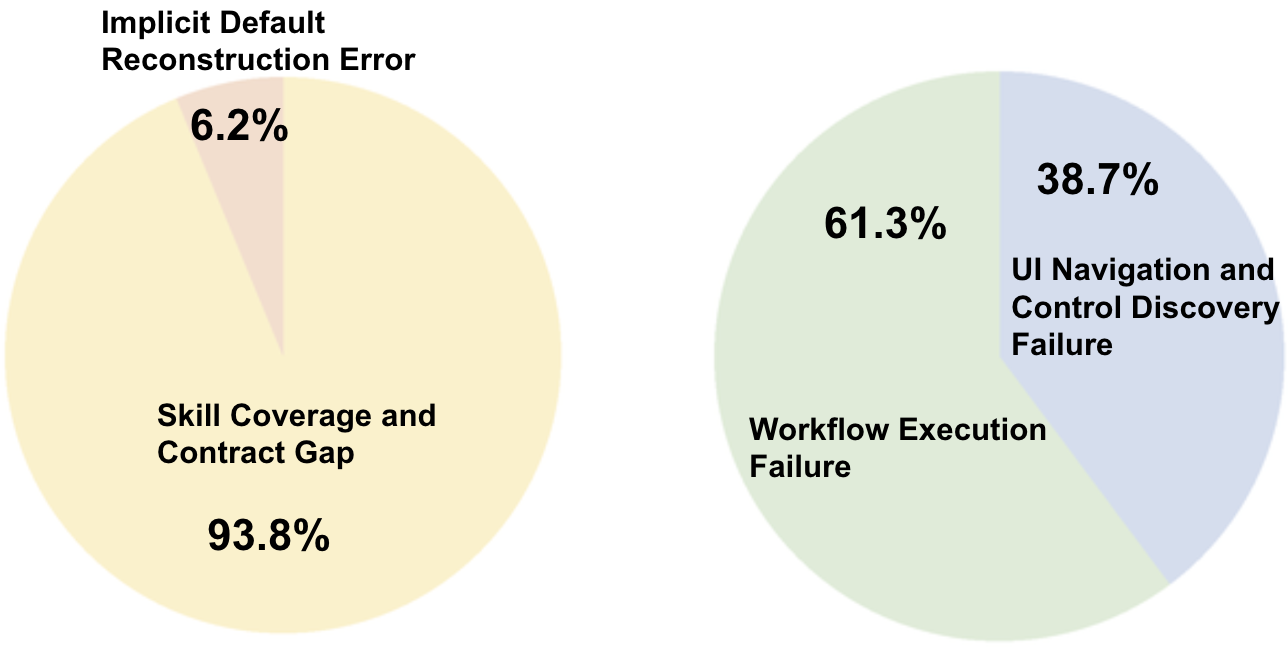}
    \caption{Distribution of primary error types for failed GUI and CLI tasks.}
    \label{fig:error-taxonomy}
\end{figure}

We manually analyze and annotate 80 randomly sampled failed trajectories per modality, drawn from failed tasks of Codex GPT-5.5 in CLI and GPT-5.4 in GUI, assigning each failed task to a single primary failure type. \mbox{Figure~\ref{fig:error-taxonomy}} summarizes the resulting distribution: CLI failures are concentrated in skill coverage and contract gaps (93.8\%), while GUI failures are split between workflow execution failures (61.3\%) and UI navigation/control discovery failures (38.7\%). The plotted distribution uses collapsed primary labels: self-checking GUI failures are assigned to the navigation or workflow failure they co-occur with, and unobservable application semantics is assigned to skill coverage and contract gaps when the missing semantics is not exposed by the skill interface.

\subsection{CLI Error Taxonomy}

CLI failures mainly arise when the skill-defined command interface does not expose, preserve, or serialize verifier-relevant application state. We identify three recurring errors:

\paragraph{Skill Coverage and Contract Gap.} In the CLI modality, application capabilities are mediated through skills and harnesses, so the executable action boundary is effectively determined by the skill contract. As illustrated by several trajectories, failures arise both when required operations are not exposed by the skill library and when documented skill behavior diverges from the underlying implementation. As a result, agents may follow the prescribed workflow while still failing final verifier checks on the produced artifacts.

\paragraph{Implicit Default Reconstruction Error.} In the CLI modality, agents must explicitly reconstruct application-specific identity conventions that GUI users often inherit implicitly from the interface, such as default object names, identifier assignment rules, and the distinction between internal names and user-facing labels. Across several trajectories, failures arise when task specifications omit information that would normally be supplied through GUI defaults or interface conventions. As a result, agents may create structurally correct objects while assigning incorrect identifiers or default values, leading to artifacts that do not satisfy the intended application state.

\paragraph{Unobservable Application Semantics.} In the CLI modality, agents are limited to the information exposed through commands, scripts, and skill interfaces. When critical application semantics are not explicitly exposed, agents often compensate by hallucinating plausible default behaviors or hidden application rules. Across several trajectories, this leads agents to produce outputs that appear structurally correct while still violating implicit application semantics that were incorrectly inferred from incomplete information.

\begin{table*}[!t]
\centering
\small
\setlength{\tabcolsep}{3.2pt}
\resizebox{\textwidth}{!}{
\begin{tabular}{l*{12}{c}>{\columncolor{SummaryColGray}}c>{\columncolor{SummaryColGray}}c}
\toprule
\textbf{Setting}
& \textbf{Vis.} & \textbf{Aud.} & \textbf{Know.} & \textbf{CAD.} & \textbf{Gra.} & \textbf{Doc.} & \textbf{Vid.} & \textbf{Sheet.} & \textbf{Slide.} & \textbf{Comm.} & \textbf{Game} & \textbf{Web}
& \textbf{Avg.} & \textbf{Time (s)} \\
\midrule
GUI GPT-5.4 & \textcolor{black}{47.1} & \textcolor{black}{73.5} & \textcolor{black}{42.9} & \textcolor{black}{46.9} & \textcolor{black}{\textbf{51.2}} & \textcolor{black}{51.3} & \textcolor{black}{60.5} & \textcolor{black}{\textbf{61.1}} & \textcolor{black}{\textbf{78.1}} & \textcolor{black}{70.0} & \textcolor{black}{84.2} & \textcolor{black}{\textbf{88.2}} & \textcolor{black}{59.1} & \textcolor{black}{455.8} \\
Original Skill  & \textcolor{black}{54.9} & \textcolor{black}{42.9} & \textcolor{black}{22.4} & \textcolor{black}{67.3} & \textcolor{black}{9.8} & \textcolor{black}{\textbf{64.1}} & \textcolor{black}{60.5} & \textcolor{black}{47.2} & \textcolor{black}{46.9} & \textcolor{black}{50.0} & \textcolor{black}{\textbf{100.0}} & \textcolor{black}{35.3} & \textcolor{black}{48.2} & \textcolor{black}{188.1} \\
Patched Skill & \textcolor{black}{\textbf{66.7}} & \textcolor{black}{\textbf{81.6}} & \textcolor{black}{\textbf{81.6}} & \textcolor{black}{\textbf{73.5}} & \textcolor{black}{48.8} & \textcolor{black}{56.4} & \textcolor{black}{\textbf{86.8}} & \textcolor{black}{41.7} & \textcolor{black}{65.6} & \textcolor{black}{\textbf{95.0}} & \textcolor{black}{\textbf{100.0}} & \textcolor{black}{35.3} & \textcolor{black}{\textbf{69.3}} & \textcolor{black}{\textbf{162.6}} \\
\midrule
Relative Change & \textcolor{black}{$\uparrow$21.5\%} & \textcolor{black}{$\uparrow$90.2\%} & \textcolor{black}{$\uparrow$264.3\%} & \textcolor{black}{$\uparrow$9.2\%} & \textcolor{black}{$\uparrow$398.0\%} & \textcolor{black}{$\downarrow$12.0\%} & \textcolor{black}{$\uparrow$43.5\%} & \textcolor{black}{$\downarrow$11.7\%} & \textcolor{black}{$\uparrow$39.9\%} & \textcolor{black}{$\uparrow$90.0\%} & \textcolor{black}{0.0\%} & \textcolor{black}{0.0\%} & \textcolor{black}{$\uparrow$43.8\%} & \textcolor{black}{$\downarrow$13.6\%} \\
\bottomrule
\end{tabular}
}
\caption{Diagnostic effect of verifier-guided skill patching for Codex GPT-5.5, with GUI GPT-5.4 from Table~\ref{tab:domain-results} included only as a reference point. Workflow-category columns report full pass rate, ``Avg.'' reports the overall full pass rate across all 440 tasks, and ``Time (s)'' reports average wall-clock time per task. The best value among the three setting rows is highlighted in \textbf{bold}; for Time, lower is better. Relative change is computed between the patched-skill and original-skill CLI settings; downward arrows in pass-rate columns indicate lower success, while the downward arrow in time indicates faster execution. The patched-skill setting uses verifier information during repair and should be interpreted as a coverage-controlled upper bound rather than an out-of-the-box CLI baseline.}
\label{tab:skill-quality}
\end{table*}

\subsection{GUI Error Taxonomy}
For GUI agents, failures mainly arise from interaction-heavy and stateful execution through visual interfaces, menus, dialogs, and application workflows. Compared with CLI execution, GUI tasks typically require longer action trajectories and more sequential state transitions, leading to failures in control discovery, workflow completion, and final-state verification. We identify three recurring GUI error types:

\paragraph{UI Navigation and Control Discovery Failure.} In the GUI modality, agents must visually discover and navigate application-specific control surfaces such as menus, tabs, dialogs, and hidden settings. Across several trajectories, failures occur when agents cannot reliably locate the correct interaction path or bind actions to the intended control surface. As a result, verifier-relevant settings or objects often remain unchanged, missing, or attached to the wrong interface state.

\paragraph{Workflow Execution Failure.} In the GUI modality, agents must complete tasks through application-specific sequences of interface operations involving menus, dialogs, confirmations, and ordered actions. Across several trajectories, failures occur when agents cannot infer the correct step-by-step workflow required to realize the target state change, such as applying operations in the wrong order, missing required confirmation steps, or terminating the workflow prematurely. As a result, verifier-relevant settings, objects, or exported artifacts often remain incomplete, missing, or incorrectly configured.

\paragraph{Self-Checking and Verification Gap.} In the GUI modality, agents often rely on visible interface feedback and terminate execution without explicitly verifying that the required artifacts or persisted state changes were actually produced. Across several trajectories, agents complete plausible interaction workflows and immediately declare success, but fail to perform a final verification step such as checking exported files, confirming saved state, or validating application outputs against task requirements. As a result, workflows may appear complete from the interface perspective while still failing verifier checks on the final artifact set.

\section{Further Analysis}

We next move from explaining observed failures to testing two candidate execution bottlenecks: CLI skill coverage and GUI procedural grounding.

\begin{table*}[t]
\centering
\scriptsize
\setlength{\tabcolsep}{4pt}
\begin{tabularx}{\textwidth}{@{}>{\hsize=0.8\hsize\linewidth=\hsize}X>{\hsize=1.2\hsize\linewidth=\hsize}X@{}}
\toprule
\textbf{Original task description} & \textbf{Procedure-guided description} \\
\midrule
Audacity is open with /home/user/Music/empty.aup loaded. The project currently has zero audio tracks. Create three new mono audio tracks and give each a specific name, in this exact top-to-bottom order: `Drums' (top-most track), `Bass' (middle), `Synth' (bottom). Save the project so the changes are persisted to /home/user/Music/empty.aup.
&
Audacity is open with /home/user/Music/empty.aup loaded. The project currently has zero audio tracks. Create three new mono audio tracks and give each a specific name, in this exact top-to-bottom order: \uline{1. Drums (top-most track) 2. Bass (middle) 3. Synth (bottom).} \uline{How to create each track: Use Tracks > Add New > Mono Track. A new empty mono wavetrack will appear at the bottom of the track list. Repeat three times so the project has three empty mono tracks.} \uline{How to rename each track: click the small drop-down arrow on the track's control panel, next to the default track name `Audio Track', and choose `Name...', or double-click the track name label. Type the desired name and press Enter.} \uline{Save the project with Ctrl+S so the three named tracks are persisted to /home/user/Music/empty.aup.} \\
\bottomrule
\end{tabularx}
\caption{Example of procedure-guided task rewriting for GUI execution. The procedure-guided description preserves the same target state and verifier, but exposes the application workflow needed to reach that state through the GUI.}
\label{tab:gui-grounding-example}
\end{table*}

\subsection{How Much of the CLI Gap Comes from Skill Coverage?}

The error analysis suggests that many CLI failures are caused by incomplete or incorrect skill interfaces.
We therefore run a diagnostic patched-skill experiment to estimate how much CLI performance is recoverable if these observed skill-coverage gaps are removed.
This experiment is not intended as a new fair baseline for deployed CLI agents; rather, it is a counterfactual analysis of the original CLI setting under verifier-observed skill completeness.
To quantify and reduce this limitation, we apply a four-phase verifier-coverage pipeline to each CLI-Anything application skill folder. We first develop a detailed instruction skill that guides Codex GPT-5.5 in analyzing, extending, validating, and synchronizing CLI-Anything skills against verifier requirements. Codex GPT-5.5 is then used to execute each phase under this instruction framework: first, construct a code-level mapping between verifier checkpoints and the existing skill implementation by jointly inspecting verifier code, harness code, and skill documentation, and label each checkpoint as Pass, Partial, or Fail, where Pass indicates that the skill can reliably satisfy the corresponding verifier requirement. Then, define skill-level verifier coverage as the fraction of checkpoints labeled Pass. Under this metric, the original CLI-Anything skills achieve 37.6\% coverage. We then repair the skill paths labeled Partial or Fail and validate each application using an app-level test suite that exercises all verifier checkpoints through the skill interface. Since the repair process explicitly uses verifier information, the resulting 100\% coverage should be interpreted as verifier-observed coverage completion, not as evidence that the patched skills would automatically generalize to unseen tasks.

We then use the best-performing GUI agent as a reference to quantify how much of the CLI performance gap can be recovered by addressing incomplete skills observed during verification. Table~\ref{tab:skill-quality} shows that verifier-observed skill incompleteness explains a large fraction of the original CLI gap. Moving from the original skill setting to the patched-skill setting increases overall success rate from 48.2\% to 69.3\%, with particularly large gains in structured workflow categories such as Knowledge, Graphics Debugging, and Communication. This suggests that many CLI failures are caused by incomplete or insufficiently expressive skill interfaces rather than model limitations. In addition, patched skills reduce average execution time, indicating more efficient task execution with fewer exploratory actions. Under this diagnostic setting, CLI exceeds the strongest GUI agent overall and in several workflow categories. We do not treat this as an out-of-the-box modality comparison, because skill patching uses verifier information during repair. Instead, the result indicates that much of the original CLI deficit is recoverable when missing skill functionality is supplied. At the same time, CLI still underperforms in Spreadsheet and Web tasks, showing that skill patching does not uniformly eliminate the remaining GUI--CLI gap across workflows.

These results suggest that skill coverage is a major bottleneck for skill-mediated CLI agents, and that verifier-observed coverage gaps account for a substantial portion of the measured CLI shortfall. However, modality-specific differences persist even under verifier-guided skill patching, suggesting that interaction modality cannot be fully reduced to the availability of skill operations.

\subsection{Does Procedural Grounding Reduce GUI Execution Errors?}

We next test whether GUI failures can be reduced by making the task procedure more explicit. This diagnostic is motivated by our observation that many GUI failures do not arise from misunderstanding the final goal, but from executing it through long sequences of screen-level interactions, hidden controls, dialog confirmations, and application-specific workflows. To test procedural grounding, we compare the benchmark's modality-neutral instructions with procedure-guided descriptions, which provide explicit step-by-step workflow cues. Table~\ref{tab:gui-grounding-example} shows a concrete example of this rewriting process. We therefore construct a procedure-guided subset of 176 tasks whose successful execution primarily depends on following GUI workflows to modify application state or documents, rather than directly editing configuration files or relying on external environment effects. We then evaluate GUI agents under these procedure-guided instructions while keeping the environment, verifier, action space, and model fixed.

Table~\ref{tab:gui-grounding} shows that procedure-guided descriptions modestly improve GUI performance while reducing interaction cost. Full pass rate changes from 59.7\% to 60.2\%, average reward changes from 0.7401 to 0.7576, and average runtime decreases from 397.0s to 314.8s. These results suggest that procedural cues help agents identify more direct action paths and reduce unnecessary exploration. However, the improvement in task completion remains modest despite a reduction in execution time, indicating that workflow knowledge is only one component of GUI interaction. Even when menu paths, dialog actions, and save/export steps are made explicit, agents must still correctly ground interface elements, track state across interactions, and execute long action sequences.

\begin{table}[!t]
\centering
\small
\setlength{\tabcolsep}{6pt}
\resizebox{\linewidth}{!}{%
\begin{tabular}{lccc}
\toprule
\textbf{Setting} & \textbf{Full Pass} & \textbf{Avg. Reward} & \textbf{Time (s)} \\
\midrule
Before Grounding & \textcolor{black}{59.7\%} & \textcolor{black}{0.7401} & \textcolor{black}{397.0} \\
After Grounding & 60.2\% & 0.7576 & 314.8 \\
\midrule
Change & \textcolor{black}{$\uparrow$0.8\%} & \textcolor{black}{$\uparrow$2.4\%} & \textcolor{black}{$\downarrow$20.7\%} \\
\bottomrule
\end{tabular}
}
\caption{Effect of procedure-guided grounding on GPT-5.4 for 176 GUI application-state modification tasks. Procedure-guided descriptions preserve the same verifier and target state while adding workflow cues such as menu paths, dialog confirmations, action order, and exact object names. Relative change is measured against the original setting, and ↓ time indicates faster execution. Avg. Reward denotes the mean fraction of verifier checkpoints passed per task.}
\label{tab:gui-grounding}
\end{table}

\section{Conclusion}

By holding task goals, initial states, and final-state verifiers fixed while systematically varying the native action interface, our benchmark makes visible a distinction that is easy to miss in separate GUI and programmatic evaluations: agents do not simply execute the same task through different controls, but receive different representations of what can be done. GUI performance is strongest and more stable when the interface itself exposes the workflow, yet remains constrained by visual grounding, state tracking, and long interaction chains; making procedures explicit mostly reduces wasted exploration rather than eliminating failures. CLI performance is more variable because it depends on whether the skill layer exposes verifier-relevant state as reliable operations; verifier-guided repairs recover much of the original shortfall, but they also reveal skill construction and validation as the central scaling problem. Thus, the GUI--CLI comparison is less a ranking of interfaces than a design question for computer-use systems: robust agents need executable task structure to be made available somewhere, whether through visible workflows, verified skill interfaces, or hybrid environments that combine both.

\section*{Limitations}

Our evaluation is designed to isolate interaction modality, not to estimate the upper bound of fully unconstrained production agents. Real deployed systems may freely combine screenshots, terminals, scripts, direct file edits, database access, and application APIs. Those hybrid strategies can be highly effective, but they also make it difficult to tell whether success comes from GUI interaction, programmatic control, or bypassing the application interface altogether. We therefore restrict each agent to modality-native actions. This choice makes the GUI--CLI comparison cleaner, but it also means that our results should not be read as a claim about the best possible performance of agents with unrestricted tool access. A natural next step is to evaluate unconstrained agents separately and measure their cross-modality behavior.

The patched-skill setting is also diagnostic rather than a deployable baseline. During repair, we use verifier-observed requirements to identify missing or non-serializing skill functionality, then validate the repaired skills against those same verifier checkpoints. This coverage measure reflects whether the patched skill interface can generate verifier-readable checkpoint states in app-level validation; it does not guarantee that an autonomous agent will select and compose the right skill calls during full task execution. This setup is useful for estimating how much of the original CLI shortfall comes from skill coverage, but it does not show that the patched skills would generalize to unseen tasks, new applications, or different verifier designs. Whether skills can be constructed automatically with this level of coverage without relying on verifier-specific information remains an open question.

Finally, our failure taxonomy is meant to expose recurring patterns, not to impose perfectly separable root causes. GUI failures in particular often mix control discovery with workflow execution: an agent may fail to find the right menu because it does not know the workflow, or fail to complete the workflow because it never discovers the necessary control. We assign each sampled failure to a single primary category for analysis, but the resulting proportions should be interpreted as a coarse summary of dominant failure modes rather than as mutually exclusive causal labels.

\bibliography{anthology,custom}

\clearpage
\appendix
\section{Appendix}

\subsection{Case Studies}
\label{app:case-studies}

\paragraph{Case Study Selection.}
To further understand how interaction modality changes task execution, we inspect paired tasks where the two modalities diverge maximally. We select two groups of cases: tasks where all original-skill CLI agents pass and all GUI agents fail, and tasks where all GUI agents pass and all original-skill CLI agents fail. We use these cases to examine what each modality makes easy or difficult for the agent. These cases show that modality acts as a capability filter: CLI succeeds when the skill layer exposes verifier-relevant artifact structure as direct operations, while GUI succeeds when the application interface exposes the intended workflow more completely than the available skill layer.

Although CLI skills offer more compact and structured information than GUI screenshots, their effectiveness is strongly conditioned on the quality of the underlying skill layer. While existing CLI-anything-style skill frameworks provide a promising path toward automated skill construction, there remains a substantial gap in both the quality and coverage of the generated skills compared to what is required for robust real-world execution, raising important scalability concerns for skill-based agents.

\paragraph{Case Where CLI Succeeds and GUI Fails.}\mbox{}

\begin{tcolorbox}[
    colback=black!7.5!white,
    colframe=black!30!white,
    fontupper=\small,
    before skip=0.6em,
    after skip=0.9em,
    boxsep=5pt,
    left=5pt,
    right=5pt,
    top=5pt,
    bottom=5pt
]
\textbf{Case 1: drawio\_multipage\_microservices}\\
Create a 4-page microservices diagram at /home/user/Documents/microservices.drawio. Rename pages to: Overview, Services, Data, Deployment. Page 1 (Overview): Client, Gateway, Backend connected sequentially (2 edges). Page 2 (Services): 4 shapes labeled UserService, OrderService, PaymentService, NotificationService. Page 3 (Data): 3 shapes labeled UsersDB, OrdersDB, Cache. Page 4 (Deployment): Kubernetes and Monitoring. Save the file.
\end{tcolorbox}

CLI agents complete this task by using cli-anything-drawio to directly create and rename pages, add labeled shapes, add connectors, and inspect the resulting diagram structure. GUI agents fail because the task requires a long visual construction chain across multiple pages. In the observed GUI trajectories, agents reach the step budget after creating only part of the diagram, leaving most required pages, labels, or connectors missing from the saved artifact.

\paragraph{Case Where GUI Succeeds and CLI Fails.}\mbox{}

\begin{tcolorbox}[
    colback=black!7.5!white,
    colframe=black!30!white,
    fontupper=\small,
    before skip=0.6em,
    after skip=0.9em,
    boxsep=5pt,
    left=5pt,
    right=5pt,
    top=5pt,
    bottom=5pt
]
\textbf{Case 2: audacity\_add\_chapter\_labels}\\
Audacity is open with /home/user/Music/audiobook.aup loaded. The project contains a single wavetrack (5 seconds of narration) and no label tracks. Add a label track and place four chapter labels on it, in this order from earliest to latest: `Intro', `Chapter 1', `Chapter 2', `Outro'. Save the project so the label track and its four labels are persisted to /home/user/Music/audiobook.aup. Exact timestamps do not matter -- only that there are exactly four labels with those exact titles on one label track.
\end{tcolorbox}

GUI agents complete this task through Audacity's visible label-track workflow: create a label track, insert labels, type the required titles, and save the project. CLI agents fail because the available skill-mediated path does not reliably create the label track and label entries in the verifier-visible Audacity project state; failed runs often preserve the original project and wavetrack but leave the label-track checks unsatisfied.

\subsection{Verifier Coverage Pipeline}
\label{app:verifier-coverage-pipeline}

We audit and improve CLI-Anything skills with a four-phase verifier-coverage pipeline. The unit of analysis is a verifier checkpoint rather than a task. Coverage is judged by the final persisted artifact or application state that the verifier actually reads. Each phase is manually inspected and approved before the next phase begins.

\begin{tcolorbox}[
    colback=black!7.5!white,
    colframe=black!30!white,
    fontupper=\small,
    boxsep=5pt,
    left=5pt,
    right=5pt,
    top=5pt,
    bottom=5pt
]
\textbf{Four-phase verifier-coverage pipeline}
\begin{enumerate}[leftmargin=*]
    \item \textbf{Verifier-to-skill mapping.} For each application, inspect the verifier implementation, verifier tests when available, skill documentation, and CLI harness code. For every public verifier checkpoint, record what artifact or state the verifier reads, the current skill path if one exists, and whether the skill can produce the verifier-readable final state.
    \item \textbf{Skill implementation and tests.} For each checkpoint marked Partial or Fail, add or repair the skill implementation so that it produces the verifier-compatible artifact state. Add comprehensive tests that generate outputs through the skill and validate them with the real verifier endpoints.
    \item \textbf{Coverage report.} Write an app-level README summarizing the final coverage result, validation command, pass/fail totals, changed files, and any residual limitations.
    \item \textbf{Skill documentation update.} Update the app's SKILL.md from the final implemented CLI code, so that the documented commands and options match the capabilities agents can actually invoke.
\end{enumerate}
\end{tcolorbox}

Each verifier checkpoint is assigned one of three labels. \textit{Pass} means the current skill can stably generate the final state read by the verifier. \textit{Partial} means the skill covers only a narrow subset of the checkpoint or updates intermediate state that is not serialized into the verifier-readable artifact. \textit{Fail} means no stable skill path exists. Command names or documentation claims alone do not count as evidence; the decision is based on implementation code and the verifier-readable final artifact.

After an application skill is repaired, we run an app-level comprehensive test that uses only the skill interface to generate verifier-readable artifacts and then evaluates those artifacts against every verifier checkpoint for that application. This test is required before the application is counted as fully covered.

The original-skill coverage is computed as:
\[
\frac{\#\text{Pass checkpoints before repair}}{\#\text{all verifier checkpoints}}.
\]
This gives 37.6\% coverage. The improved-skill coverage uses the same denominator after repairing and validating all \textit{Partial} and \textit{Fail} checkpoints, giving 100\%.

\subsection{Visual Design Example}
\label{app:visual-design-example}

\begin{tcolorbox}[
    colback=black!7.5!white,
    colframe=black!30!white,
    fontupper=\small,
    boxsep=5pt,
    left=5pt,
    right=5pt,
    top=5pt,
    bottom=5pt
]
\textbf{drawio\_kubernetes\_cluster}\\
Create a Kubernetes cluster architecture at /home/user/Documents/k8s\_cluster.drawio. Create 12 shapes: Control Plane, Ingress, Worker1, Worker2, Worker3, Pod1--6, ServiceMesh. Connect hierarchically: Control Plane to each Worker, each Worker to 2 Pods (Worker1 $\rightarrow$ Pod1/2, Worker2 $\rightarrow$ Pod3/4, Worker3 $\rightarrow$ Pod5/6), and Ingress to Control Plane. The diagram must have 12 vertices and at least 9 edges. Save the file.
\end{tcolorbox}

This Visual Design task is visually presented as a diagram-construction problem, but the required final state is largely structural: the artifact must contain a fixed set of vertices, specific node labels, and a minimum set of directed edges encoding the cluster hierarchy. A CLI workflow can address these properties directly through diagram-level operations such as adding labeled shapes and connecting source-target pairs, whereas a screen-only GUI workflow must place shapes, edit labels, draw connectors, and preserve the resulting diagram structure through visual interactions.

\subsection{Grounded Examples for the Error Taxonomy}
\label{app:error-taxonomy}
To support the error taxonomy in the main text, we provide grounded examples based on execution trajectories and verifier outputs. Tables~\ref{tab:cli-error-taxonomy-examples} and~\ref{tab:gui-error-taxonomy-examples} show representative cases for each failure category in CLI and GUI agents.
\begin{table*}[!t]
\centering
\scriptsize
\setlength{\tabcolsep}{3pt}
\renewcommand{\arraystretch}{1.12}
\begin{tabularx}{\textwidth}{@{}p{0.07\textwidth}p{0.17\textwidth}X X@{}}
\toprule
\textbf{Modality} & \textbf{Error taxonomy} & \textbf{Task} & \textbf{Trajectory-based explanation} \\
\midrule
CLI &
Skill coverage and contract gap &
FreeCAD is open with /home/user/Documents/parts\_catalog.FCStd. The document contains four primitive objects whose internal names are Box1, Cylinder1, Sphere1, Torus1. Do not change the internal names, but update the Label property of each object to: Box1 $\rightarrow$ StorageBox, Cylinder1 $\rightarrow$ Pipe, Sphere1 $\rightarrow$ Ball, Torus1 $\rightarrow$ Ring. Save the file. &
In freecad\_rename\_four\_objects, the agent discovered cli-anything-freecad, but invoking it on the native .FCStd file made the harness parse the file as its own JSON project format and produced a UnicodeDecodeError. It then found no stable skill command for changing labels in a native FreeCAD document and returned [INFEASIBLE]. The verifier still found the original objects, but all label checks failed. \\
\midrule
CLI &
Implicit default reconstruction error &
Add a Part::Sphere object to the document at /home/user/Documents/shapes.FCStd. Set its Radius to 15 mm and change its Label to Planet. Preserve the existing Part::Box object named Box. Save the file. &
In freecad\_add\_sphere\_to\_doc, the agent created and saved a sphere-like object, and the verifier observed objects named Box and Planet. The benchmark expected the internal FreeCAD name to remain the GUI default Sphere, with only the user-facing Label set to Planet. The checks for object Sphere, its type, radius, and label all failed because the internal name Sphere was missing. \\
\midrule
CLI &
Unobservable application semantics &
Audacity is already running. Change the Default Sample Format setting to 32-bit float. Save the preference so it persists to the configuration file. &
In audacity\_preference\_default\_sample\_format, the Audacity CLI harness did not expose global preferences, so the agent could not find a skill-mediated way to set the verifier-relevant preference. It instead inferred a plausible configuration field from the exposed state and treated `DefaultProjectSampleFormatChoice=Format32BitFloat` as the target setting. However, the requested "Default Sample Format" was actually stored as `Quality/DefaultSampleFormat=524293`, which was left unchanged. Thus, the task remained incomplete, illustrating how the agent can map exposed state to a plausible but incorrect hidden preference semantic.
 \\
\bottomrule
\end{tabularx}
\caption{Representative CLI failures supporting the error taxonomy.}
\label{tab:cli-error-taxonomy-examples}
\end{table*}

\begin{table*}[!t]
\centering
\scriptsize
\setlength{\tabcolsep}{3pt}
\renewcommand{\arraystretch}{1.12}
\begin{tabularx}{\textwidth}{@{}p{0.07\textwidth}p{0.17\textwidth}X X@{}}
\toprule
\textbf{Modality} & \textbf{Error taxonomy} & \textbf{Task} & \textbf{Trajectory-based explanation} \\
\midrule
GUI &
UI navigation and control discovery failure &
A baseline Zoom config file exists at /home/user/.config/zoomus.conf. Configure Zoom's virtual background feature so the user's background is blurred and hardware-accelerated. Update keys across two sections of /home/user/.config/zoomus.conf: Set EnableVirtualBackground to true, set VirtualBackgroundType to exactly blur, and set BlurStrength to exactly 50. Set EnableHWAccel to true. Save it in-place at the same path. Do not modify any other keys or sections. &
In zoom\_virtual\_background\_setup, the GUI agent performed only 11 screen actions: it clicked into settings-like surfaces, searched for terms such as ``hardware acceleration'' and ``blur strength'', clicked around the upper-right UI, and ended with DONE. The verifier showed that all target settings remained unchanged, including EnableVirtualBackground=false, VirtualBackgroundType=none, BlurStrength=0, and EnableHWAccel=false. \\
\midrule
GUI &
Workflow execution error &
Zotero is installed and a pre-seeded library is at /home/user/Zotero/zotero.sqlite. A PDF file is provided at /home/user/Documents/attention\_paper.pdf. Launch Zotero and wait for the library to load. Locate the existing item titled Attention Is All You Need. Attach the PDF /home/user/Documents/attention\_paper.pdf to that item as a child attachment. After attaching, the item must have at least one PDF attachment child. Leave Zotero running. &
In zotero\_gap\_attach\_pdf\_to\_item, the GUI agent executed a 34-step sequence, including right-clicking in the library, entering the PDF path, and making repeated file/menu selections. The verifier confirmed that the target item existed, but check-item-attachment failed with has\_attachment=false and attachment count 0. The agent reached the right context but did not complete the stateful attach-file workflow. \\
\midrule
GUI &
Self-checking and verification gap &
Audacity is open with /home/user/Music/podcast.wav loaded (a 12-second mono 44.1 kHz tone). Add a new label track. Create exactly three labels in that label track: at time 0.0s with title intro, at time 4.0s with title main, at time 8.0s with title outro. Save the Audacity project to /home/user/Music/podcast.aup. Export the label track to /home/user/Music/podcast\_labels.txt. Do not export the audio itself -- only the labels text file and the .aup project are required outputs. &
In audacity\_gap\_label\_track\_export, the GUI agent successfully created the label track and the three required labels at the required timestamps. The failure was the final exported artifact: /home/user/Music/podcast\_labels.txt did not exist, and the file-size check failed with ``File not found.'' The agent interacted with an export dialog and stopped with DONE, but did not verify that the text export was actually written. \\
\bottomrule
\end{tabularx}
\caption{Representative GUI failures supporting the error taxonomy.}
\label{tab:gui-error-taxonomy-examples}
\end{table*}

\subsection{Prompts}
\label{app:prompts}
To enforce modality-specific execution constraints in our benchmark, we design prompts that restrict agents to either skill-based CLI actions or screen-level GUI interactions. This ensures that both modalities operate within their native action spaces without cross-modality leakage.
\begin{figure*}[t]
\centering
\begin{tcolorbox}[
    colback=black!7.5!white,
    colframe=black!30!white,
    fontupper=\scriptsize\ttfamily,
    fonttitle=\scriptsize,
    boxsep=5pt,
    left=5pt,
    right=5pt,
    top=5pt,
    bottom=5pt
]
\textless SYSTEM\_CAPABILITY\textgreater\\
* You are utilising an Ubuntu virtual machine (Docker container) using x86\_64 architecture with internet access.\\
* You can feel free to install Ubuntu applications with your bash tool. Use curl instead of wget.\\
* DO NOT ask users for clarification during task execution. DO NOT stop to request more information from users. Always take action using available tools.\\
* When using bash commands that are expected to output very large quantities of text, redirect into a tmp file and use Read or grep to confirm output.\\
* TASK FEASIBILITY: You can declare a task infeasible at any point during execution. If you determine that a task cannot be completed, output exactly "[INFEASIBLE]" (including the square brackets) anywhere in your response.\\
* Home directory of this Ubuntu system is '/home/user'.\\
* If you need a password for sudo, the password of the computer is 'user'.\\
* The current date is \{date\}.\\[0.4em]
\textless IMPORTANT\_CAPABILITIES\textgreater\\
* You have access to CLI-anything skills for various applications installed in the system.\\
* The system includes desktop applications for office work, graphics editing, development, media playback, music notation, and more.\\
* Available CLI-anything skills can be discovered and used as needed for different target applications.\\[0.4em]
* You also have access to standard tools with strict limits:\\
\hspace*{1em}- Bash: Use for read-only inspection, skill discovery, invoking CLI-anything/application commands, and verification.\\
\hspace*{1em}- Read: Inspect files when needed.\\[0.4em]
\textless CLI\_ANYTHING\_ONLY\_POLICY\textgreater\\
You must solve the task by using CLI-anything skills for the target application.\\[0.4em]
Hard requirements:\\
* First discover the relevant CLI-anything skill(s) for the application and use those skill-provided commands/workflows as the primary and mandatory way to complete the task.\\
* All task-result changes must be performed through CLI-anything skill workflows or application-level commands explicitly recommended by those skills.\\
* You must not directly modify task files, project files, application config files, databases, document archives, or other target artifacts with generic file-editing methods.\\
* Do not use Python, sed, awk, perl, node scripts, shell redirection, heredocs, Write/Edit tools, or manual archive/database editing to change the target artifact.\\
* Bash is allowed only for read-only inspection, skill discovery, invoking CLI-anything/application commands, and verification. It must not be used to directly rewrite target files.\\
* Python is allowed only for read-only inspection if absolutely necessary. It must never write, patch, serialize, save, or mutate task artifacts.\\
* If the task cannot be completed through CLI-anything skills or skill-approved application commands, output exactly "[INFEASIBLE]" instead of bypassing the restriction with direct file edits.\\
\textless/CLI\_ANYTHING\_ONLY\_POLICY\textgreater\\[0.4em]
\textless TASK\_EXECUTION\_APPROACH\textgreater\\
Use this order:\\
1. Identify the target application and discover its CLI-anything skill.\\
2. Follow the skill's documented workflow to perform the requested changes.\\
3. Use read-only commands to inspect results when needed.\\
4. If no CLI-anything path exists, declare "[INFEASIBLE]".\\[0.4em]
Direct file mutation is not an acceptable fallback.\\
\textless/TASK\_EXECUTION\_APPROACH\textgreater\\
\textless/IMPORTANT\_CAPABILITIES\textgreater\\
\textless/SYSTEM\_CAPABILITY\textgreater
\end{tcolorbox}
\caption{CLI agent system prompt.}
\label{fig:cli-system-prompt}
\end{figure*}

\begin{figure*}[t]
\centering
\begin{tcolorbox}[
    colback=black!7.5!white,
    colframe=black!30!white,
    fontupper=\scriptsize\ttfamily,
    fonttitle=\scriptsize,
    boxsep=5pt,
    left=5pt,
    right=5pt,
    top=5pt,
    bottom=5pt
]
\textless GUI\_SCREEN\_ONLY\_POLICY\textgreater\\
Hard requirements:\\
* All task-result changes must be made through the target application's visible GUI.\\
* Do not directly rewrite task files on disk or mutate their underlying data using Python, shell commands, scripts, automation APIs, databases, archives, config files, or external utilities.\\
* Editing content through the target application's own GUI is allowed.\\
* Opening a terminal, REPL, scripting console, developer console, or macro/script editor to execute code or commands is not allowed, even if accessed through the GUI.\\
\textless/GUI\_SCREEN\_ONLY\_POLICY\textgreater
\end{tcolorbox}
\caption{GUI agent screen-only policy.}
\label{fig:gui-screen-only-policy}
\end{figure*}

\end{document}